\documentclass[11pt]{article}
\usepackage{rotating}
\usepackage{acl2015}
\usepackage{times}
\usepackage{url}
\usepackage{latexsym}
\usepackage{tabularx}
\usepackage{makecell}
\usepackage{hyperref}
\usepackage{multirow}
\usepackage{multicol}
\usepackage{booktabs}
\usepackage{appendix}
\usepackage{tcolorbox}
\usepackage{colortbl}
\usepackage{listings}
\usepackage{arydshln}
\usepackage{CJKutf8}
\usepackage{amsmath}
\usepackage[hang,flushmargin]{footmisc} 
\usepackage[T1]{fontenc}
\usepackage{lipsum}
\usepackage{graphicx}

\usepackage{fancyhdr}
\usepackage{caption}
\tcbuselibrary{breakable}

\title{REAL: Benchmarking Abilities of Large Language Models for Housing Transactions and Services}

\author{Kexin Zhu , Yang Han\footnotemark[1] \\
Beike Inc., Beijing, China  \\
\texttt{\{zhukexin008,hanyang030\}@ke.com}}

\begin{document}
\maketitle
\begin{abstract}
The development of large language models (LLMs) has greatly promoted the progress of chatbot in multiple fields. There is an urgent need to evaluate whether LLMs can play the role of agent in housing transactions and services as well as humans.
We present \textbf{R}eal \textbf{E}state \textbf{A}gent \textbf{L}arge Language Model Evaluation (REAL), the first evaluation suite designed to assess the abilities of LLMs in the field of housing transactions and services. REAL comprises 5,316 high-quality evaluation entries across 4 topics: memory, comprehension, reasoning and hallucination. 
All these entries are organized as 14 categories to assess whether LLMs have the knowledge and ability in housing transactions and services scenario. 
Additionally, the REAL is used to evaluate the performance of most advanced LLMs. The experiment results indicate that LLMs still have significant room for improvement to be applied in the real estate field.
\end{abstract}


\renewcommand{\thefootnote}{\fnsymbol{footnote}}
\footnotetext[1]{ Corresponding author.}
\renewcommand{\thefootnote}{\arabic{footnote}}

\section{Introduction}


\begin{figure}[htb]
\centering
\includegraphics[width=\linewidth]{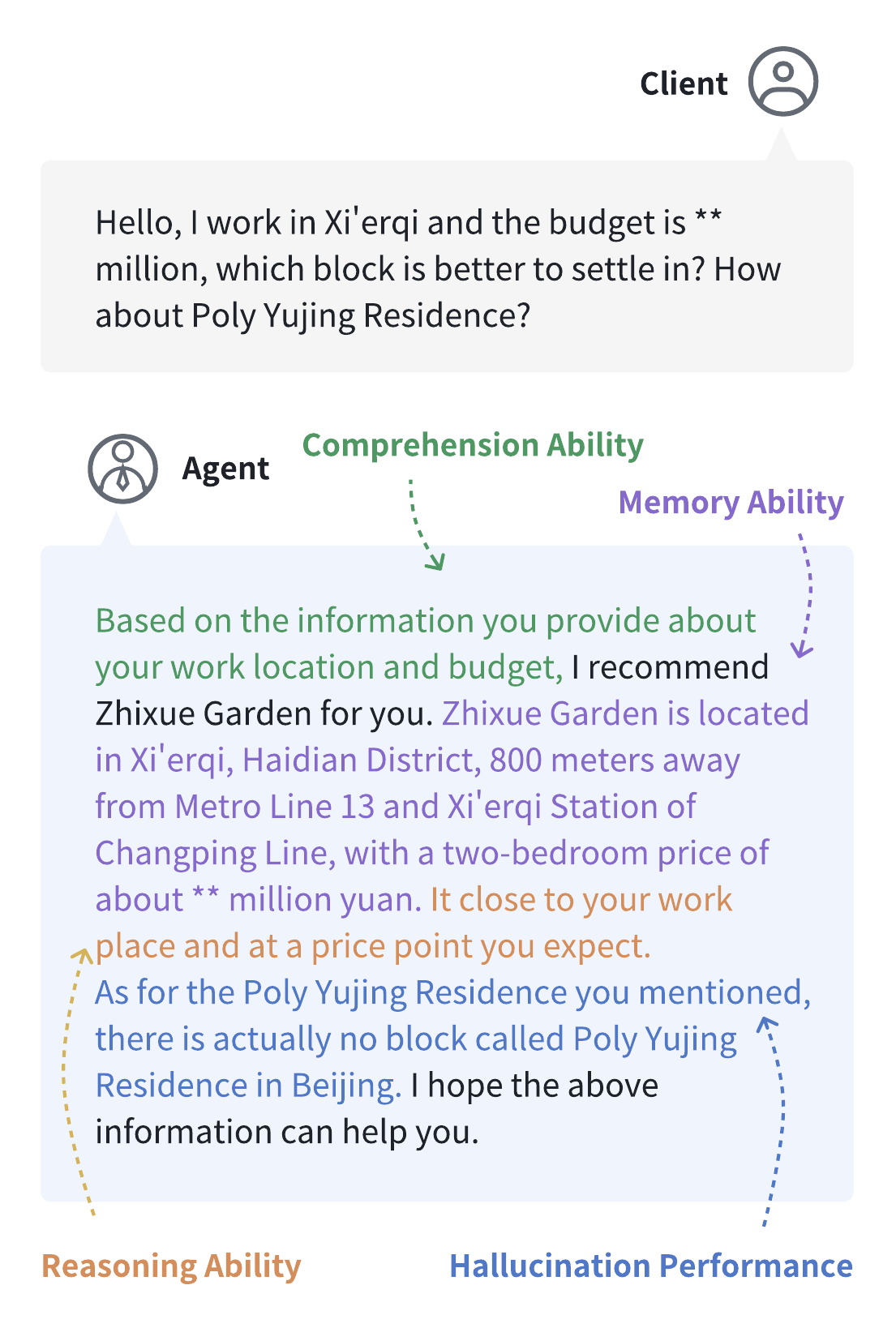}
\caption{\label{fig:scenarios} Abilities that chatbot should have in housing transactions and services scenario.}
\end{figure}

Nowadays, the LLMs are applied in online service scenarios, such as legal advice, financial advice, disease diagnosis and after-sales service etc. 
The chat between customers and real estate consultant is a typical scenario. 
An expert should have extensive knowledge of housing, excellent understanding, reasoning skills and proficient business skills as shown in Figure \ref{fig:scenarios}. 
In order to evaluate whether LLMs have the same abilities as experts in this field,
the specific benchmarks are needed.

Recently, many benchmarks for LLMs have been proposed. However, most of these benchmarks only focus on the general capability of LLMs and cannot reflect the performance in some specific fields.
On the other hand, building benchmarks for specific fields requires professional knowledge from domain expertise. This makes the process of building the benchmark extremely time-consuming and labor-intensive.

\begin{figure*}[htb]
\centering
\includegraphics[width=\linewidth]{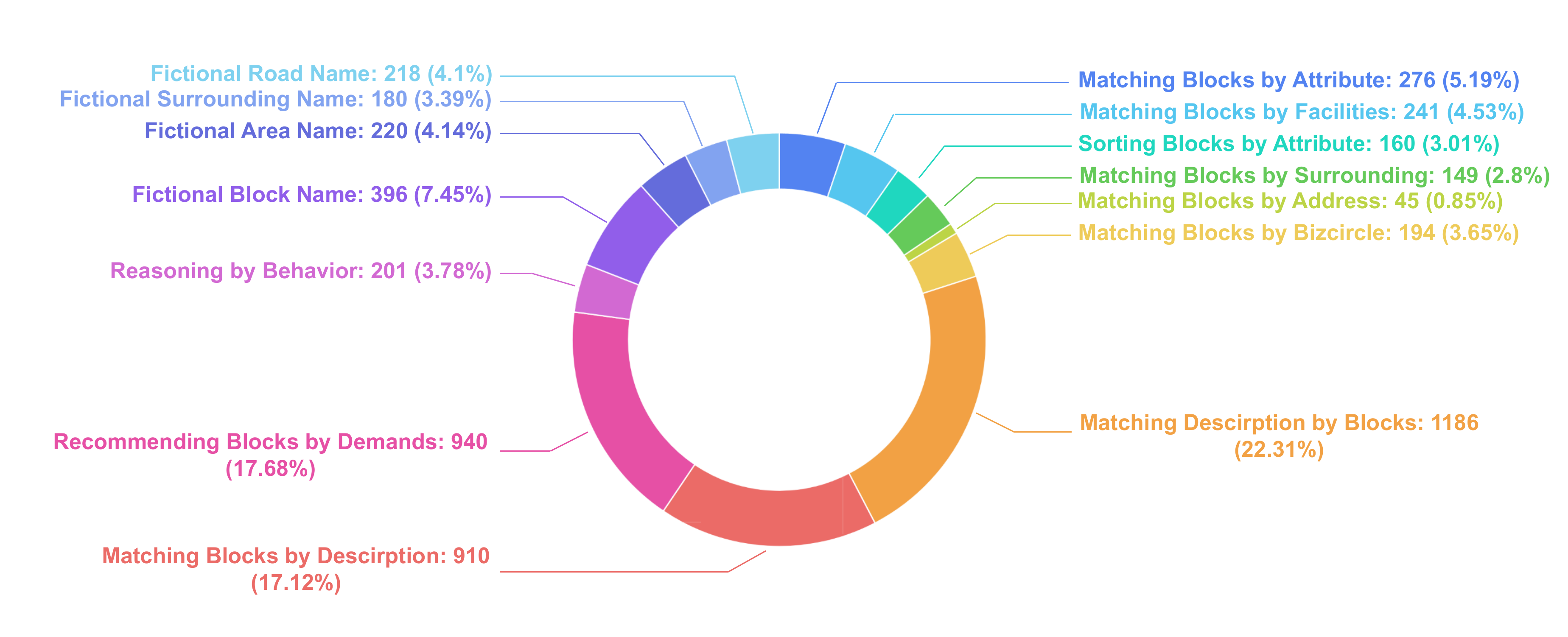}
\caption{\label{fig:distribution}Overview of the REAL.}
\end{figure*}

To evaluate whether LLMs have the abilities of memory, comprehension and reasoning in housing transactions and services,
we propose the REAL benchmark.
The REAL contains 5,316 high-quality questions across four topics.
All questions and the corresponding answers are carefully designed by experts in the real estate field. 
We use REAL to evaluates the capabilities of mainstream LLMs for housing transactions and services. 

The main contributions are as follows:
\begin{itemize}
    \item We propose a professional benchmark specifically for the housing transactions and services. The benchmark contains three topics about \textbf{Memory}, \textbf{Comprehension} and \textbf{Reasoning}.
    There are 4,302 high-quality evaluation entries across these three topics.
    \item We construct a topic about \textbf{Hallucination} to evaluate the performance of LLMs in the field of housing transactions and services. There are 1,014 entries for the evaluation of hallucination.
    \item We evaluate the performance of the mainstream LLMs in the real estate field, and provide a reference for application and optimization of models. All prompts and data examples are listed in Appendix \ref{app:a} \ref{app:b} \ref{app:c}.
\end{itemize}

\section{Related Works}
To evaluate the performance of LLMs, different kinds of benchmarks have emerged, such as C-Eval \cite{NEURIPS2023_c6ec1844}, CMMLU \cite{li-etal-2024-cmmlu}, OpenbookQA \cite{mihaylov-etal-2018-suit}, etc. Most of these benchmarks focus on the general capabilities or reasoning abilities of LLMs. For example, The C-Eval focuses on assessing the world knowledge of LLMs. It covers more than 13,900 multiple-choice questions across 52 disciplines such as STEM and social science. Each discipline has a certain level of difficulty. 
The CMMLU is specifically used to evaluate the knowledge and reasoning ability of LLMs in the Chinese context. It covers 67 different topics ranging from basic disciplines to advanced professional levels. The OpenbookQA contains difficulties such as multi-step reasoning, common sense knowledge and text understanding. It uses a form similar to open-book exam to evaluate the performance of comprehension about a certain topic. These benchmarks have many evaluation dimensions and involve many basic world knowledge. 
However, these benchmarks cannot cover the professional knowledge of many disciplines.
Using these benchmarks to evaluate the performance of LLMs in specific fields is insufficient.

\begin{figure*}[htb]
\centering
\includegraphics[width=0.98\linewidth]{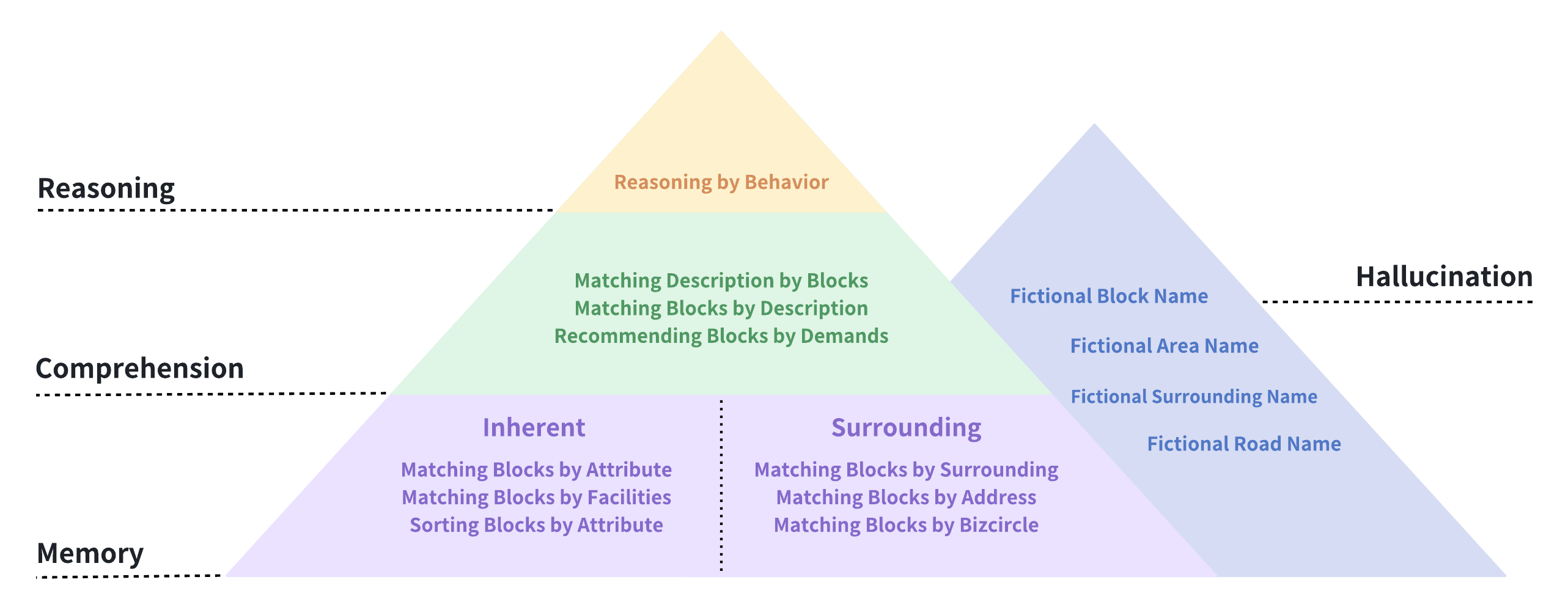}
\caption{\label{fig:design}Pyramid Structure of REAL.}
\end{figure*}

With the application of LLMs across different fields, researchers have also proposed benchmarks for some specific fields. For example, the OpenFinData \cite{openfindata}  designs 19 tasks based on actual financial business scenarios. The OpenFinData focuses on the evaluation in the financial field.
The Fin-Eva \cite{fineva} contains more than 13,000 questions, covering multiple financial scenarios such as wealth management, insurance and investment research, as well as the knowledge of professional financial subject disciplines. 
The LawBench \cite{fei-etal-2024-lawbench} simulates three dimensions of judicial cognition and evaluates the legal capabilities of LLMs about the China's legal system. 
The ChemBench \cite{zhang2024chemllmchemicallargelanguage} evaluates the chemical knowledge of LLMs through 4,100 multiple-choice questions about chemical molecules and chemical reactions.
The SecBench \cite{secbench} covers various computer security certificate examinations and network security event theoretical questions. It provides effective evaluation capabilities for the research and development of security. 
The Flames \cite{huang-etal-2024-flames} is highly adversarial and is dedicated to evaluate the performance of models in three dimensions such as fairness, morality and legality. 
The establishment of these benchmarks for specific fields requires experienced experts.

As of now, there is still a gap in benchmarks for evaluating the capabilities of LLMs in the scenario of housing transaction and services.

\section{REAL}

\subsection{Overview}
The REAL evaluates the performance of LLMs for housing transactions and services from 3 topics: memory, comprehension and reasoning ability.
At the same time, we also construct a topic about hallucination.
Therefore, the REAL measures the performance of LLMs from 4 topics comprehensively.
All these 4 topics are further divided into 14 categories.

Figure \ref{fig:distribution} shows the distribution of various categories in the REAL. The REAL contains a total of 5,316 high-quality questions. All questions and corresponding answers are checked by annotator. It is worth noting that the scope of all entries in REAL is limited to Beijing and does not involve other cities just for now.

\subsection{Evaluation Topic}
We summarize various abilities that an expert should possess and divide them into topics.
In addition, we notice the hallucination problem of LLMs in online service scenarios.
We specially design a topic to evaluate the hallucination performance of LLMs in housing transactions and services.

As shown in Figure \ref{fig:design}, the REAL includes 4 topics: memory, comprehension, reasoning and hallucination. The first 3 topics focus on the ability in real estate brokerage. Each entry contains a question, four options and additional explanatory notes. 
The last topic involves the evaluation of hallucination. Each entry in hallucination topic contains a query and a premise. Examples are listed in Appendix \ref{app:a}.

\textbf{Memory Topic} Memory topic concerns if LLMs have the knowledge about blocks. The knowledge focuses on the inherent attributes of blocks (the ratio of grassland, the amount of parking spots, etc.) and the surrounding of the blocks (the distance between block and some facilities). 
All entries in this topic are further divided into 6 categories. 
For example, LLMs are required to match the block with the specific attribute or bizcircle. 
Every entry in this topic is multiple-choice question and only one answer is correct.
The LLMs' performance is judged by percent of correct answers.

\begin{figure*}
\centering
\includegraphics[width=\linewidth]{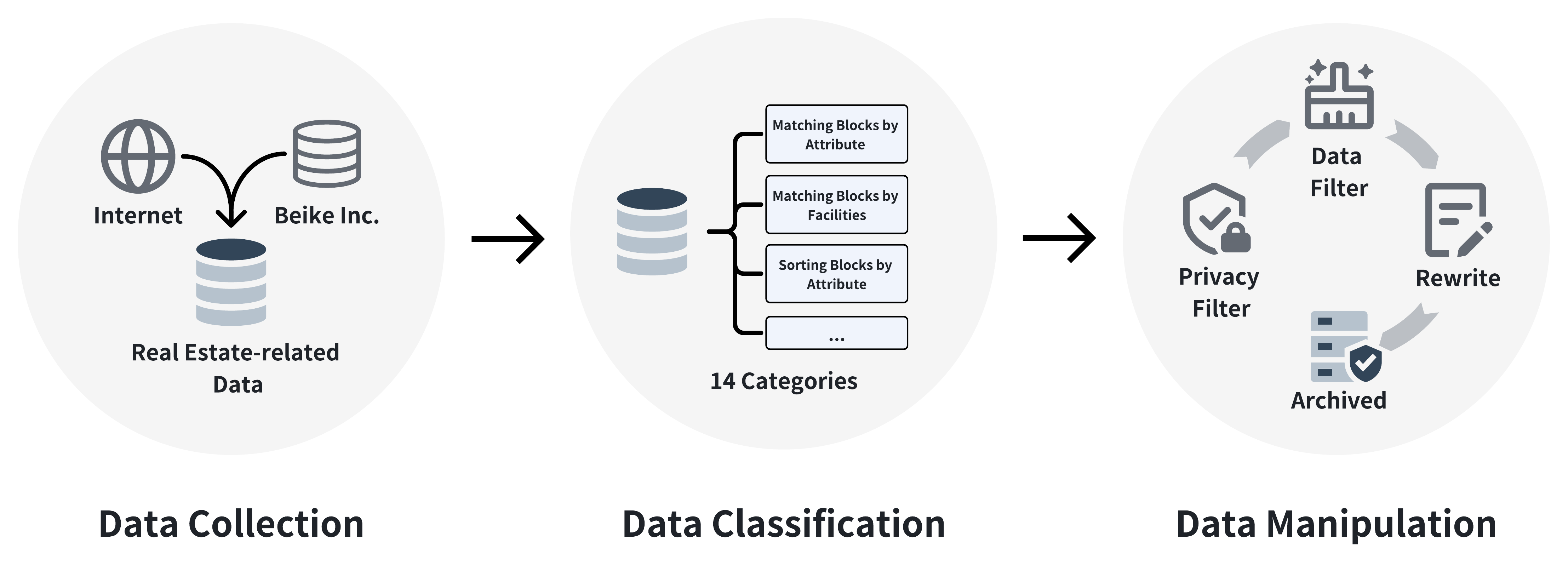}
\caption{\label{fig:process}REAL Data Pipeline.}
\end{figure*}

\textbf{Comprehension Topic} Comprehension topic concern whether LLMs have the ability to understand natural language descriptions. 
For example, one of the tasks is to check if the LLMs can associate the description with the most consistent block entity. This topic contains 3 categories. Details are listed in \ref{fig:design}.
Entries in this topics are also multiple-choice.
The performance is judged by whether the answers selected by LLMs are correct.

\textbf{Reasoning Topic} Reasoning ability is an advanced and complex skill and is built upon knowledge and comprehension skills. Reasoning topic contains tasks that reasoning what type of house the client is likely to purchase based on their past behavior data. The data of behavior involve browsing and clicking information on websites of BeiKe.

\textbf{Hallucination Topic} Hallucination topic is specially designed for LLMs. A human agent usually doesn't make up non-existent entities to the customer, and a LLM should behave like this as an agent. Hallucination is evaluated based on the natural language inference (NLI) methods \cite{maynez-etal-2020-faithfulness} \cite{manakul-etal-2023-selfcheckgpt}. Specifically, we use queries with fictional names to get model's responses. 
Then, we use another model with strong capabilities as a judge to determine whether there is a conflict between the premises and the model's responses. The hallucination topics include 4 categories and all these entries is a pair of query and premise. The response of the judge model is used as the evaluation criterion. All prompts for the judge model are listed in Appendix \ref{app:b}.

\subsection{Data Pipeline}

In order to construct a comprehensive and reliable benchmark, we design a standard data pipeline. All entries in the REAL are produced by this pipeline strictly, as shown in Figure \ref{fig:process}. The details are shown in Appendix \ref{app:a}

\textbf{Data Collection} All real estate-related data are collected from Beike corporation.
Additionally, we also collect queries from clients about residential issues from the Internet. All these data is collected in legal and compliant ways.

\textbf{Data Classification}  Data are divided into 14 categories to assess the model's performance on specific tasks. Ten of the categories simulate various scenarios that real estate agents may encounter in their daily work. 
The other 4 categories are designed to test the hallucinatory performance of the model in residential-related conversations. We design the format, rules and prompts for each category in advance. 
We offer several examples for small-scale verification to ensure that the solution is feasible before large-scale manufacture. The details of each category are shown in Appendix \ref{app:a}.

\textbf{Data Manipulation} Instead of using these data directly, we use a variety of methods to filter out privacy information. These methods include character matching, semantic detection and security model detection. After all these manipulation, we confirm that the data does not contain any sensitive information, such as name, telephone number, etc.
Subsequently, we inspect all these data to filter out unreasonable entries.
For example, the ratio of grassland in certain block is too high or too low. And the area of the house that do not meet the number of rooms obviously.
Finally, there is still some problems with data such as typos or abbreviations. Therefore we rewrite the collected entries to a certain extent. At the same time, We ensure that the original meaning of the sentence is preserved.
Prompts used in the process of data manipulation are listed in Appendix \ref{app:b}.

\begin{table*}[htb]
\centering
\begin{center}
\caption{\label{tab:data}Statistics of memory, comprehension, and reasoning topic.}
\renewcommand\arraystretch{1}
\begin{tabular}{llc}
\toprule \noalign{\vskip 0.5mm}
\textbf{Topic}                     & \textbf{Category}                & \textbf{Quantity} \\ \noalign{\vskip 0.5mm}\midrule
\multirow{6}{*}{Memory}        & Matching Blocks by Attribute    & 276                \\
                                     & Matching Blocks by Facilities     & 241                \\
                                     & Sorting Blocks by Attribute      & 160                \\
                                     & Matching Blocks by Surrounding & 149                \\
                                     & Matching Blocks by Address        & 45                 \\
                                     & Matching Blocks by Bizcircle    & 194                \\ \noalign{\vskip 1mm}\cdashline{1-3}[1pt/2pt]\noalign{\vskip 1mm}
\multirow{3}{*}{Comprehension} & Matching Description by Blocks  & 1186               \\
                                     & Matching Blocks by Description  & 910                \\
                                     & Recommending Blocks by Demands     & 940                \\ \noalign{\vskip 0.5mm}\cdashline{1-3}[1pt/2pt]\noalign{\vskip 1mm}
Reasoning                     & Reasoning by Behavior        & 201                \\ \noalign{\vskip 1mm}\cdashline{1-3}[1pt/2pt]\noalign{\vskip 1mm}
\multicolumn{2}{c}{Total}                                                 & 4302               \\ \bottomrule
\end{tabular}
\end{center}
\end{table*}

\begin{table}[htb]
\centering
\begin{center}
\caption{\label{tab:datahallu}Statistics of hallucination topic.}
\renewcommand\arraystretch{1}
\begin{tabular}{lc}
\toprule \noalign{\vskip 0.5mm}
\textbf{Category}          & \textbf{Quantity} \\ \noalign{\vskip 0.5mm}\midrule
Fictional Block Name       & 396               \\
Fictional Area Name        & 220               \\
Fictional Surrounding Name & 180               \\
Fictional Road Name        & 218               \\ \noalign{\vskip 1mm}\cdashline{1-2}[1pt/2pt]\noalign{\vskip 1mm}
Total                      & 1014              \\ \bottomrule
\end{tabular}
\end{center}
\end{table}

\subsection{Benchmark Construction}

\begin{table*}[htb]
\centering
\begin{center}
\caption{\label{tab:models}Evaluated Models.}
\renewcommand\arraystretch{1}
\begin{tabular}{lllcc}
\toprule \noalign{\vskip 0.5mm}
\textbf{Model}         & \textbf{Creator} & \textbf{Release} & \textbf{Parameters}  & \textbf{Assess} \\ \noalign{\vskip 0.5mm}\midrule
GPT-4o-2024-05-13      & OpenAI           & 2024/5/13        & \textit{undisclosed} & API             \\
GPT-4o-2024-08-06      & OpenAI           & 2024/8/6         & \textit{undisclosed} & API             \\
GPT-4-Turbo-2024-04-09 & OpenAI           & 2024/4/9         & \textit{undisclosed} & API             \\
Internal Model V1 & BeiKe          & -                & -                    & Weights             \\
Internal Model V2 & BeiKe          & -                & -                    & Weights             \\
Qwen2-72B-Instruct     & Alibaba          & 2024/6/7         & 72B                  & Weights         \\
Qwen2.5-72B-Instruct   & Alibaba          & 2024/9/18        & 72B                  & Weights         \\
DeepSeek-V2-Lite-Chat  & Deepseek         & 2024/5/16        & 16B                  & Weights         \\
GLM-4-9B-Chat          & ZhipuAI          & 2024/6/5         & 9B                   & Weights         \\
Baichuan2-13B-Chat     & Baichuan         & 2023/9/6         & 13B                  & Weights         \\
Yi-1.5-34B-Chat        & 01-AI            & 2024/5/13        & 34B                  & Weights         \\ \bottomrule
\end{tabular}
\end{center}
\end{table*}

\textbf{Memory Evaluation} 
For the topic of memory, We build two sets: P set and Q set.
The P set is a collection of physical properties and the Q set is a collection of styles of question.
For example, the P set contains various attributes of the block such as ratio of grassland and number of parking spots. The Q set contains various styles of question such as sorting certain items in ascending order or sorting some properties in descending order.
Every question is generated by combining entries from P set and Q set randomly.
For example, the property "ratio of grassland" and the style "sorting certain items in ascending order" are randomly sampled from P and Q set respectively.
Then the stem of question is "which is the correct order to sort the block by the ratio of grassland in ascending order? "
Subsequently, we extract four blocks from the factual knowledge data and sort these blocks in ascending order by the ratio of grassland to get the correct answer. The other three wrong ranking orders are created as confusion options. 
It is important to note that the P and Q sets are different for each category. 
For example, the Q set only contains two styles for the category that matching block by bizcircle.
While the Q set has three kinds of styles for category that matching block by surroundings.
To preserve diversity, we also build a set T which contains multiple query templates. Template is randomly selected from T set and will be filled with items sampled from P and Q set.

\textbf{Comprehension Evaluation} 
For the topic of comprehension, We sample blocks and its description from the data. The stem of the question is either the name of a certain block or the description of a certain block.
The task is matching the correct description based on the name of a block or vice versa. The confusion options are extracted from the other blocks or descriptions.
For example, the question is that "which of the following four options is the correct description of the given block? ". More examples are shown in Appendix \ref{app:a}.
In particular, for the category of recommending block by clients' requirements, we invite experienced real estate agents to build the options. Besides, the reasons why these options meet the clients' requirements or not are also annotated by these real estate agents.

\textbf{Reasoning Evaluation} 
For the topic of reasoning, the task is to infer which block the client is most likely to buy based on their past behavior and profile.
LLMs is used to summarize users' portrait into an easy-to-read natural language form as the context of the question. 
The query is that "based on the above context information, which block the client is most likely to buy? ".
The correct answer is based on the fact and the other three confusions are blocks that are mentioned in context but not purchased by the client.
We also build a set H which contains blocks different with the correct block. If there is not enough blocks in the context, blocks from set H are used.    

\textbf{Hallucination Evaluation} 
For the category of hallucination, we make up some non-existent items (names of block, surrounding, etc.) by LLMs or manually. Then we filter out the items that are duplicated or too similar with the names in reality.
Subsequently, we use these non-existent items to replace the real items in the query to obtain the hallucination evaluation questions. The original queries are also maintained as positive sample.
Each question is also accompanied by premise that if the items is real or non-existent.
For example, if block A is real, then the premise is that block A exists. 
And if block B is fictional, the premise is that block B does not exist. 
During the evaluation process, the premise are used as a reference for the judge model.
In order to increase the difficulty, we make the style of the fictional names diverse and close to reality. The details information are listed in Appendix \ref{app:a}.

\textbf{Quality Verification} All questions and options are verified by annotators manually before archived.
The items to be checked include sensitive information, correctness and accuracy. 
Entries that fail the verification will be destroyed to ensure that the benchmark is accurate and compliant.
After the inspection, all these entries will be archived in the standard JSON format by topics and categories. 
The 10 categories for memory, comprehension, and reasoning topic are shown in Table \ref{tab:data}. The 4 categories for hallucination topic are shown in Table \ref{tab:datahallu}.

\section{Experiments}

\begin{table*}[htb]
\centering
\begin{center}
\caption{\label{tab:abilityresult}Average accuracy (\%) in memory, comprehension and reasoning.}
\renewcommand\arraystretch{1}
\begin{tabular}{lcccc}
\toprule \noalign{\vskip 0.5mm}
\textbf{Model}         & \textbf{Memory }$\uparrow$ & \textbf{Comprehension }$\uparrow$ & \textbf{Reasoning }$\uparrow$ & \textbf{Average }$\uparrow$ \\ \noalign{\vskip 0.5mm}\midrule
Internal Model V2        & \textbf{78.22}  & 67.73                  & 44.78              & \textbf{71.73}   \\
Internal Model V1         & 65.32           & \textbf{75.73}         & 45.77              & 66.49            \\
GPT-4o-2024-05-13      & 45.01           & 62.70                  & 48.26              & 50.64            \\
GPT-4o-2024-08-06      & 45.26           & 61.07                  & 49.75              & 50.45            \\
Qwen2-72B-Instruct     & 42.27           & 61.59                  & 47.76              & 48.62            \\
Qwen2.5-72B-Instruct   & 41.98           & 60.94                  & 47.76              & 48.25            \\
GPT-4-Turbo-2024-04-09 & 36.90           & 50.77                  & \textbf{51.24}     & 42.50            \\
Yi-1.5-34B-Chat        & 33.63           & 50.91                  & 46.27              & 40.07            \\
GLM-4-9B-Chat          & 33.05           & 48.06                  & 47.26              & 38.98            \\
Baichuan2-13B-Chat     & 32.61           & 43.35                  & 40.30              & 36.60            \\
DeepSeek-V2-Lite-Chat  & 33.96           & 40.51                  & 32.34              & 35.77            \\ \noalign{\vskip 1mm}\cdashline{1-5}[1pt/2pt]\noalign{\vskip 1mm}
Random                 & 25.00              & 25.00                     & 25.00                 & 25.00              \\ \bottomrule
\end{tabular}
\end{center}
\end{table*}

\subsection{Setup}
We focus on giving an overview of LLMs' capabilities by the REAL. 
We select most advanced LLMs in Chinese context. All these models are shown in Table \ref{tab:models}. 
All open-source models are deployed and invoked locally, such as Qwen series models \cite{yang2024qwen2technicalreport} \cite{qwen2.5}, DeepSeek \cite{deepseekai2024deepseekv2strongeconomicalefficient}, GLM \cite{glm2024chatglmfamilylargelanguage}, Baichuan2 \cite{yang2023baichuan2openlargescale}, Yi \cite{ai2024yiopenfoundationmodels} and our internal models.
Models of GPT-4 \cite{openai2024gpt4technicalreport} series are invoked through API.

All these models are evaluated in zero-shot settings and the temperature is set to 0.1.
The precision of all open-source models is set to BF16. 
For memory, comprehension, and reasoning evaluations, we require all the models to produce the uppercase letter (A, B, C, or D) corresponding to the selected option. Otherwise we select the first uppercase letter in the output text as the answer. 
We calculate the accuracy of the model's responses across all 10 categories.
The average accuracy of all categories are used as a measure of the model's ability in the residential field. 

For hallucination evaluation, we use GPT-4 as the judge model. 
The hallucination rate $r_{hallu}$ is calculated as formula \ref{hallu}:
\begin{equation}\label{hallu}
    r_{hallu} = \frac{n_{hallu}}{N}
\end{equation}
$n_{hallu}$ is the number of hallucination and $N$ is the total number of questions in every category. Additionally, we also calculate the hallucination rate with and without fictional names separately to demonstrate the effect of fictitious names.

\begin{table*}[htb]
\centering
\begin{center}
\caption{\label{tab:halluresult}Average hallucination rate (\%) in hallucination evaluation.}
\renewcommand\arraystretch{1}
\begin{tabular}{lccc}
\toprule \noalign{\vskip 0.5mm}
\textbf{Model}         & \textbf{\begin{tabular}[c]{@{}c@{}}Average $\downarrow$ \\ (w fictional entity)\end{tabular}} & \textbf{\begin{tabular}[c]{@{}c@{}}Average $\downarrow$ \\ (w/o fictional entity)\end{tabular}} & \textbf{\begin{tabular}[c]{@{}c@{}}Average $\downarrow$ \\ (total)\end{tabular}} \\ \noalign{\vskip 0.5mm}\midrule
Internal Model V1         & \textbf{60.91}                                                                    & 0.61                                                                                 & \textbf{30.76}                                                      \\
Internal Model V2        & 69.17                                                                             & 0.66                                                                                 & 34.91                                                               \\
Qwen2-72B-Instruct     & 72.72                                                                             & 0.23                                                                                 & 36.47                                                               \\
GPT-4-Turbo-2024-04-09 & 71.96                                                                             & 1.52                                                                                 & 36.74                                                               \\
GPT-4o-2024-08-06   & 81.93                                                                             & \textbf{0.00}                                                                                 & 40.96                                                               \\
Qwen2.5-72B-Instruct      & 81.94                                                                             & 1.70                                                                        & 41.82                                                               \\
Baichuan2-13B-Chat     & 84.24                                                                             & 1.55                                                                                 & 42.90                                                               \\
GPT-4o-2024-05-13      & 86.40                                                                             & 0.36                                                                                 & 43.38                                                               \\
DeepSeek-V2-Lite-Chat  & 86.72                                                                            & 0.36                                                                                 & 43.54                                                               \\
Yi-1.5-34B-Chat          & 84.74                                                                             & 5.58                                                                                 & 45.16                                                               \\
GLM-4-9B-Chat        & 93.00                                                                             & 1.60                                                                                & 47.30                                                               \\ \bottomrule
\end{tabular}
\end{center}
\end{table*}

\begin{table*}[htb]
\centering
\begin{center}
\caption{\label{tab:corrresult}Pearson Correlation and accuracy (\%) in agreement evaluation.}
\renewcommand\arraystretch{1}
\begin{tabular}{lcccc}
\toprule \noalign{\vskip 0.5mm}
\textbf{Test Model}   & \textbf{Judge X} & \textbf{Judge Y} & \textbf{Correlation} & \textbf{Accuracy} \\ \noalign{\vskip 0.5mm}\midrule
Qwen2.5-72B-Instruct  & GPT-4            & Annotator            & 0.9796               & 99.00             \\
GPT-4o-2024-08-06     & GPT-4            & Annotator            & 0.9730               & 98.67             \\
Internal Model V2       & GPT-4            & Annotator            & 0.9579               & 98.00             \\
DeepSeek-V2-Lite-Chat & GPT-4            & Annotator            & 0.9598               & 98.00             \\
Yi-1.5-34B-Chat       & GPT-4            & Annotator            & 0.9063               & 95.33             \\ \bottomrule
\end{tabular}
\end{center}
\end{table*}

To verify the effectiveness of the judge model (and the prompts), we also conduct the agreement evaluation. 
We choose five different models include GPT, Qwen2.5, DeepSeek, Yi and the internal models. 
Subsequently, we randomly sample 300 questions from the hallucination dataset. Therefore, we obtain 1,500 pairs of queries and responses totally. The evaluation results are produced by the judge model and annotators parallelly.
For agreement evaluation, two scores are reported: Pearson Correlation and Accuracy.
For Pearson Correlation, We refer to the method in AlignBench \cite{liu-etal-2024-alignbench}. We regard the results of "without hallucination" and "with hallucination" as binary values of 0 and 1. And the results of human annotator and judge model constitute two vectors, respectively. 
Pearson Correlation $r$ is calculated by:
\begin{equation}\label{pearson}
    r = \frac{\sum{}(\boldsymbol{x} - m_x)(\boldsymbol{y} - m_y)}{\sqrt{\sum{}(\boldsymbol{x} - m_x)^2\sum{}(\boldsymbol{y} - m_y)^2}}
\end{equation}
where $\boldsymbol{x}$ and $\boldsymbol{y}$ represent judge result vectors of Judge X (model) and Judge Y (annotator), $m_x$ and $m_y$ represent the mean of $\boldsymbol{x}$ and $\boldsymbol{y}$. For accuracy, we take the results of Judge Y as ground truth and check whether the results of Judge X is correct or not. Accuracy $Acc$ is calculated by:
\begin{equation}\label{acc}
    Acc = \frac{n_{corr}}{N}
\end{equation}
where $n_{corr}$ represents the number of correct results produced by Judge X and $N$ represents the total number of results.



\subsection{Memory, Comprehension and Reasoning Evaluation Results}

All the results are shown in Table \ref{tab:abilityresult}.
Our internal model v2 achieve the best accuracy of 71.73\%. The GPT-4o achieve the best accuracy excluding our internal models. However the average accuracy is just around 50\%, indicating that 
LLMs still have significant room for improvement to be applied in the residential field directly.
In the memory evaluation, our internal models are significantly ahead of other models because of the internal data. 
In the comprehension evaluation, the REAL can distinguish all these models effectively. 
In the reasoning evaluation, we note that the gap between these models is smaller.
But the accuracy of all models are below 50\% in reasoning topic.
This phenomenon indicates that the reasoning in the residential field is still difficult for LLMs.

\subsection{Hallucination Evaluation Results}
Table \ref{tab:halluresult} shows the results of hallucination evaluation.
The results indicate that fictional names can easily lead to the generation of hallucinatory responses in conversations related to the residential field. 
When the queries contain a fictional name, the model tends to make a introduction to the fictional name rather than suspecting or correcting it. 
The results also indicate that even if the name is real, the models have a certain probability of believing that the name is fictional. That is also hallucination.

\subsection{Agreement Evaluation Results}

Table \ref{tab:corrresult} shows the results of agreement evaluation. Pearson Correlation  coefficients are all higher than 0.9. It indicates that there is a clear positive relationship between the results of Judge X and Judge Y. 
All accuracy are above 95\% in Table \ref{tab:corrresult}. 
Therefore, the result of judge model (and prompts) used in our hallucination evaluation is consistent with human judgment.

\section{Conclusion}
We propose REAL benchmark, which dedicate to evaluate the performance of LLMs for housing transactions and services. 
We design three topics including memory, comprehension and reasoning to reflect LLMs' ability.
Additionally we design an hallucination benchmark to assess LLMs' performance in the residential field. 
Through the evaluation of mainstream models by the REAL, we note that LLMs still have significant room for improvement to be applied in the real estate field.

Up to now, the REAL only focus on the information and knowledge related to Beijing. We will continue to expand and optimize the benchmark in the follow-up works and expand the scope to other cities.

\section{Acknowledgement}

We would like to thank the KE Team members for the support and beneficial discussions. We would also like to especially thank Zijun Guan, Ji Zhang, Junlong Yuan, and Yiduo Liu for their valuable contributions to the data pipeline and benchmark construction. Their knowledge and practical experience in housing transactions and services greatly supported the quality of our REAL benchmark.

\bibliographystyle{acl}
\bibliography{acl2015}

\onecolumn
\newpage
\appendix

\section{\label{app:a}Examples of Data}
Appendix A shows the introductions, formats and examples of 14 categories across 4 topics.

\subsection{Matching Blocks by Attribute}

This category examines the model's memory of the intrinsic attributes of the blocks and the simple comparison between the attributes. The form of the question is to directly ask the specific value of the inherent attributes of the block, such as the ratio of grassland, the amount of parking spots, etc.

\begin{tcolorbox}[title = {Example}, breakable]
\begin{CJK}{UTF8}{gbsn}
\textbf{问题  }
以下四个小区中，绿化率最高的是？

\textbf{选项  }
A. 北京气象局宿舍
B. 鑫顺苑
C. 幸福东区
D. 龙湖好望山

\textbf{答案  }
D
\end{CJK}
\tcblower
\textbf{Question    }
Among the following four blocks, which has the highest ratio of grassland ?

\textbf{Options    }
A. Dormitory of Beijing Meteorological Bureau.
B. Xin Shun Yuan.
C. Happiness East District.
D. Longhu Good Hope Mountain.

\textbf{Answer    }
D
\end{tcolorbox}

\subsection{Matching Blocks by Surroundings}

This category examines the model's knowledge of the facilities and distances around the blocks. The format is to select from four blosks with or without some kind of facilities within a certain distance.

\begin{tcolorbox}[title = {Example}, breakable]
\begin{CJK}{UTF8}{gbsn}
\textbf{问题  }
请从以下选项中选出一个1000米内有小学的小区：

\textbf{选项  }
A. 上庄家园
B. 华润八号院
C. 明天生活馆
D. 绿地大兴启航国际

\textbf{答案  }
D
\end{CJK}
\tcblower
\textbf{Question    }
Please choose a block with a primary school within 1000 meters from the following options:

\textbf{Options    }
A. Shangzhuang Homestead.
B. China Resources No. 8.
C. Tomorrow Life Hall.
D. Greenland Daxing set sail international.

\textbf{Answer    }
D
\end{tcolorbox}

\subsection{Matching Blocks by Address}

This category examines the model's knowledge of the address in which the block is located. The question is in the form of a given range, and from the four options, select the right block that is or is not located in that range.

\begin{tcolorbox}[title = {Example}, breakable]
\begin{CJK}{UTF8}{gbsn}
\textbf{问题  }
以下哪个小区不位于四至五环之间？

\textbf{选项  }
A. 酒仙桥三街坊
B. 远见名苑二期
C. 万科金阳
D. 大溪地

\textbf{答案  }
B
\end{CJK}
\tcblower
\textbf{Question    }
Which of the following blocks is not located between the 4th and 5th Ring Roads?

\textbf{Options    }
A. Jiuxianqiao 3 Street.
B. Vision Garden Phase II.
C. Vanke Jinyang.
D. Tahiti.

\textbf{Answer    }
B
\end{tcolorbox}

\subsection{Matching Blocks by Bizcircle}

This category examines the model's knowledge of the bizcircle in which the block is located. The question is in the form of a given bizcricle name, and from the four options, select the right block name that is located or not located in that bizcircle.

\begin{tcolorbox}[title = {Example}, breakable]
\begin{CJK}{UTF8}{gbsn}
\textbf{问题  }
以下哪个小区位于百子湾商圈？

\textbf{选项  }
A. 百合园  
B. 玉林东里三区
C. 金泰先锋南区
D. 昆仑域

\textbf{答案  }
C
\end{CJK}
\tcblower
\textbf{Question    }
Which of the following blocks is located in Baiziwan Bizcircle?

\textbf{Options }
A. Lily Garden.
B. Yulin Dongli District 3.
C. Jintai Pioneer South District.
D. Kunlun Domain.

\textbf{Answer  }
C
\end{tcolorbox}

\subsection{Matching Description by Blocks}

This category mainly examines the model's use of block knowledge and understanding of natural language form block descriptions. The form of the question is given a block, and the description that best fits the community is selected from the four descriptions.

\begin{tcolorbox}[title = {Example}, breakable]
\begin{CJK}{UTF8}{gbsn}
\textbf{问题  }
关于大柳树甲17号院平房小区的描述正确的是？

\textbf{选项  }

A. 小区周边有多个购物中心，包括京泰娇阳富硒团购超市和精品超市。

B. 小区的绿化率达到35\%，能够一定程度上缓解环境压力。

C. 小区建于2000年后，建筑结构为钢筋混凝土结构。

D. 小区周边有多个地铁站，交通便利性极高。

\textbf{答案  }
B
\end{CJK}
\tcblower
\textbf{Question    }
Which of the following options is correct about the bungalow block in Daliushujia No. 17 Courtyard?

\textbf{Options    }

A. There are a number of shopping malls around this block, including Jingtai Jiaoyang Selenium-rich group purchase supermarket and boutique supermarket.

B. The greening rate of this block reaches 35\%, which can alleviate the environmental pressure to a certain extent.

C. This block was built after 2000, and the building structure is a reinforced concrete structure.

D. There are multiple subway stations around this block, and the transportation convenience is extremely high.

\textbf{Answer    }
B
\end{tcolorbox}

\subsection{Matching Blocks by Description}

This category examines the understanding and memory of the model. The form is given a description, and the block name that best matches the description is selected from the four block names.

\begin{tcolorbox}[title = {Example}, breakable]
\begin{CJK}{UTF8}{gbsn}
\textbf{问题  }
小区的地理位置非常优越，位于北京市西城区德胜门区域，紧邻北二环和京藏高速，交通便利。小区周边有多条公交线路，最近的公交站点安德路西口距离小区仅50米，涵盖27路、380路、夜36路等多条线路。地铁方面，距离鼓楼大街地铁站仅900米，该站提供2号线和8号线的换乘服务，进一步提高了出行的便捷性。商业配套方面，小区内设有美廉美超市，周边还有新华百货和多个菜市场，满足居民的日常购物需求。医疗方面，小区附近有多家医院，如火箭军医院和德胜门中医院，提供完善的医疗服务。教育资源丰富，小区周边有多所幼儿园、小学和中学，如北京第二实验小学和北师大二附中。环境方面，小区周围有多个公园，如北滨河公园和人定湖公园，环境优美，适合居住。 以下哪一个小区与上述描述最匹配？

\textbf{选项  }
A. 昊文温泉家园
B. 安德馨居
C. 首开幸福广场
D. 向军北里

\textbf{答案  }
B
\end{CJK}
\tcblower
\textbf{Question    }
The geographical location of the block is very superior, located in the Deshengmen area of Xicheng District, Beijing, close to the North Second Ring Road and Beijing-Tibet Expressway, with convenient transportation. There are many bus lines around the block, and the nearest bus station, Ande Road West Exit, is only 50 meters away from the block, covering 27 Road, 380 Road, Night 36 Road and other lines. In terms of subway, it is only 900 meters away from the Gulou Street subway station, which provides transfer services for lines 2 and 8, further improving the convenience of travel. In terms of commercial facilities, there is a Meilianmei supermarket in the block, and there are also Xinhua Department Store and a number of vegetable markets in the vicinity to meet the daily shopping needs of residents. In terms of medical care, there are a number of hospitals in the vicinity of the block, such as the Rocket Army Hospital and the Deshengmen Hospital of Traditional Chinese Medicine, which provide comprehensive medical services. There are many kindergartens, primary schools and middle schools around the block, such as Beijing No. 2 Experimental Primary School and No. 2 High School Affiliated to Beijing Normal University. In terms of environment, there are many parks around the block, such as North Riverside Park and Rendinghu Park, which have a beautiful environment and are suitable for living. Which of the following blocks best matches the above description?

\textbf{Options    }
A. Haowen Hot Spring Home.
B. Ander Xinju.
C. Shoukai Happiness Square.
D. Xiang Jun Beili.

\textbf{Answer    }
B
\end{tcolorbox}

\subsection{Recommending Blocks by Demands}

This category of questions simulates the consultation scenario between clients and real estate agents, and comprehensively examines the model's memory of block knowledge and understanding of client needs. This part of the question is rewritten by the user's query on the social platform, which can reflect the real user needs, and the options and answers are all manually generated by experienced real estate agents. The question is in the form of a statement of client needs, and the model selects the most suitable block from the four blocks.

\begin{tcolorbox}[title = {Example}, breakable]
\begin{CJK}{UTF8}{gbsn}
\textbf{问题  }
您好，可以咨询一下吗？我目前打算出售通州的大方居两限房和蓟门桥的一室一厅，预计可以拿到700多万，然后再贷款买第二套房。总价控制在900万以内，您能推荐一下海淀的二手小区吗？我想买一个小三居，目前看上了东王庄。

\textbf{选项  }
A.光熙家园二期
B.九台庄园
C.东王庄
D.芍药居20号院

\textbf{答案  }
C

\textbf{解释（由标注人员撰写，仅供人工分析参考，不输入模型）  }

A,光熙家园二期,是2008年建成，小区设有独立园林景观设计。小区内配套有室内健身馆、游泳馆，半场篮球场、小区内做了人车分流设计，安全，十分适合有居住品质需求的客户入住,两居室的面积都在120平米以上，总价都在1150万以上,属于朝阳区西坝河，不属于海淀区，B,九台庄园,小区是封闭式小区，别墅区环境优美，绿树成荫，小区都是别墅，密度低，小区处处都是绿地，占地面积大，经过改造以后，小区路面都重新修整了，市政水，天然气，集中供暖，适合居住，属于别墅，总价已经超预算了，也不属于海淀区，C,东王庄,小区是公房社区，属于成熟社区，在1992年到2001年建成，本小区2008年被评为海淀十大文明社区，小区共38栋楼，三居室（73-96平米），总价在720-900万，物业管理比较完善，2019年年底市政给小区改造的外墙加保温、坡顶新换的瓦、做的保温层,D,芍药居20号院,小区是2000年建成的纯商品房社区，有独立的物业管理，小区有自行安装的电动车车棚，且有小孩子玩耍的小公园，滑梯秋千，小区有老年活动区域，共3栋楼，1号楼是3梯8户，共20层，2和3号楼是3梯10户，共24层，两居室（109-119平米），总价在730-800万左右，三居室（142-150平米），总价在880-1000万左右，小学是人大附朝阳学校的划片范围，属于朝阳区芍药居，不属于海淀区，故正确答案是C。
\end{CJK}
\tcblower
\textbf{Question    }
Hello, can I consult you? I currently plan to sell the two-limit house in Dafangju in Tongzhou and the one-bedroom apartment in Jimen Bridge, and I am expected to get more than 7 million, and then take out a loan to buy a second house. The total price is controlled within 9 million, can you recommend the second-hand blocks in Haidian? I want to buy a small three-bedroom and currently have my eye on Dongwangzhuang.

\textbf{Options    }
A. Guangxi Homes Phase II.
B. Jiutai Manor.
C. Dongwangzhuang.
D. Shaoyaoju No. 20.

\textbf{Answer    }
C

\textbf{Explanation (Explanations are written by annotators and are only for human analysis and do not input into the model.)  }

A, Guangxi Home Phase II, was completed in 2008, the block has an independent landscape design. There are indoor gymnasiums, swimming pools, half-court basketball courts, and the block has made a people-vehicle diversion design, safe, very suitable for clients with living quality needs, the area of the two-bedroom is more than 120 square meters, and the total price is more than 11.5 million, which belongs to Xiba River, Chaoyang District, does not belong to Haidian District, B, Jiutai Manor, the block is a closed block, the villa area has a beautiful environment, green trees, the block is a villa, the density is low, the block is full of green space, covers a large area, after transformation, The pavement of the block has been re-repaired, municipal water, natural gas, central heating, suitable for living, belongs to the villa, the total price has exceeded the budget, and does not belong to Haidian District, C, Dongwangzhuang, the block is a public housing block, belongs to a mature block, built from 1992 to 2001, the block was rated as one of the top ten civilized blocks in Haidian in 2008, the block has a total of 38 buildings, three bedrooms (73-96 square meters), the total price is 7.20-9 million, and the property management is relatively perfect, At the end of 2019, the municipal government added thermal insulation to the exterior wall of the block, the new tile on the slope roof, and the insulation layer, D, Shaoyaoju No. 20, the block is a pure commercial housing block built in 2000, with independent property management, the block has its own installed electric car carport, and there is a small park for children to play, slide swings, the block has an elderly activity area, a total of 3 buildings, Building 1 is 3 ladders and 8 households, a total of 20 floors, Buildings 2 and 3 are 3 ladders and 10 households, a total of 24 floors, two bedrooms (109-119 square meters), The total price is about 7.3 million to 8 million, the three-bedroom (142-150 square meters), the total price is about 880-10 million, the primary school is the zoning range of Chaoyang School attached to the National People's Congress, which belongs to Shaoyaoju in Chaoyang District, not Haidian District, so the correct answer is C.
\end{tcolorbox}

\subsection{Reasoning by Behavior}

This category gives a complete segment of the user's behavior (deleting the final deal information) as the context, and the model predicts the deal result based on the behavior, and selects the final deal from the four listings. This type of question is more difficult, and can comprehensively examine the ability of model knowledge application, reading comprehension, logical reasoning, predictive analysis, etc., at the same time, the results also have a certain degree of uncertainty.

\subsection{Fictional Block Name}

This category is used to detect model hallucinations, and to check whether the content generated by the model is reliable by simulating the client's real query with a fictional block name. The questions in this category include positive examples (no fictional block names injected) and negative examples (injected fictional block names).

\begin{tcolorbox}[title = {Example}, breakable]
\begin{CJK}{UTF8}{gbsn}
\textbf{问题  }
请问您能帮我分析一下保利御景华庭这个小区怎么样吗？

\textbf{虚构小区名  }
保利御景华庭

\textbf{前提  }
北京市不存在“保利御景华庭”。
\end{CJK}
\tcblower
\textbf{Question    }
Can you help me analyze the block Poly Yujing Residence?

\textbf{Fictional block name    }
Poly Yujing Residence.

\textbf{Premise    }
"Poly Yujing Residence" does not exist in Beijing.
\end{tcolorbox}

\subsection{Fictional Area Name}

This category is used to detect model hallucinations, and to test whether the content generated by the model is reliable by simulating the real query of the client and adding the fictional area names (district names, school district names, business district names, etc.) to the query. The category consists of positive examples (no fictional area names injected) and negative examples (injected fictional area names).

\begin{tcolorbox}[title = {Example}, breakable]
\begin{CJK}{UTF8}{gbsn}
\textbf{问题  }
我在瑞光门工作，预算首付大约150万左右，想购买一个户型好的两居室。对学区没有特别要求，不想要老旧的小区，希望通勤方便，居住舒适，并且抗跌性好的小区。请问您有什么推荐吗？

\textbf{虚构区域名  }
瑞光门

\textbf{前提  }
北京市不存在“瑞光门”。
\end{CJK}
\tcblower
\textbf{Question    }
I work in Ruiguangmen, with a down payment of about 1.5 million, and I want to buy a good two-bedroom apartment. There are no special requirements for the school district, and I don't want an old community, but I want a community that is convenient for commuting, comfortable to live in, and has good resistance to falls. Do you have any recommendations?

\textbf{Fictional area name    }
Ruiguangmen.

\textbf{Premise    }
"Ruiguangmen" does not exist in Beijing.
\end{tcolorbox}

\subsection{Fictional Surrounding Name}

This category is used to detect model hallucinations, and to test the reliability of the model's generated content by simulating the client's real query and adding fake names of surrounding facilities (school names, hospital names, shopping mall names, park names, etc.) to the query. There are positive examples (no fictional facility names injected) and negative examples (fictional facility names injected).

\begin{tcolorbox}[title = {Example}, breakable]
\begin{CJK}{UTF8}{gbsn}
\textbf{问题  }
请问老公通勤到朝阳，又想离健德医院通州院区近，应该买哪个小区呢？

\textbf{虚构周边名  }
健德医院通州院区

\textbf{前提  }
北京市不存在“健德医院通州院区”。
\end{CJK}
\tcblower
\textbf{Question    }
May I ask which blosk should we buy if my husband commutes to Chaoyang and wants to be close to the Tongzhou branch of Jiande Hospital?

\textbf{Fictional surrounding name    }
Tongzhou branch of Jiande Hospital.

\textbf{Premise    }
"Tongzhou branch of Jiande Hospital" does not exist in Beijing.
\end{tcolorbox}

\subsection{Fictional Traffic Name}

This category is used to detect model hallucinations, and to test whether the model generates reliable content by simulating the client's real query and adding fake traffic names (subway line names, subway station names, bus line names, bus station names, road names, ring names, railway station names, airport names, etc.) to the query. There are positive examples (no fictional traffic names injected) and negative examples (fictional traffic names injected).

\begin{tcolorbox}[title = {Example}, breakable]
\begin{CJK}{UTF8}{gbsn}
\textbf{问题  }
您好，我的工作地点在白虎门地铁站，周围辐射通勤时长在40分钟内的地区都可以接受。请问在这些区域内，有没有整租两居或者一居的房源，预算在4000元以内可以租到吗？

\textbf{虚构交通名  }
白虎门地铁站

\textbf{前提  }
北京市不存在“白虎门地铁站”。
\end{CJK}
\tcblower
\textbf{Question    }
Hello, my place of work is in Baihumen subway station, and the surrounding area with a radiating commute time of 40 minutes is acceptable. In these areas, are there any two-bedroom or one-bedroom properties that can be rented within 4,000 yuan?

\textbf{Fictional Road name    }
Baihumen subway station.

\textbf{Premise    }
"Baihumen subway station" does not exist in Beijing.
\end{tcolorbox}

\section{\label{app:b} Prompts for Data Construction}

Appendix B shows the prompts used in dataset construction.

\subsection{Query Rewrite}

\begin{tcolorbox}[title = {Prompt}, breakable]
\begin{CJK}{UTF8}{gbsn}
以下是一些询问推荐哪个小区的句子，请在不改变原意的情况下，将以下句子改写成客户询问房产经纪人的表达形式，直接输出改写后的句子：

\textbf{\{input\_content\}}
\end{CJK}
\tcblower
The following are some sentences that ask which block to recommend, without changing the original meaning, please rewrite the following sentences into the form of the customer asking the real estate agent, and directly output the rewritten sentence:

\textbf{\{input\_content\}}
\end{tcolorbox}

\subsection{Fictional Block Name Production (General)}

\begin{tcolorbox}[title = {Prompt}, breakable]
\begin{CJK}{UTF8}{gbsn}
参考样例：\textbf{\{input\_content\}}

以上是一些北京的小区名，请你模仿这些小区名再虚构至少20个小区名，要求不能与北京现实中存在的小区名重复。输出前请再检查一遍确保你输出的小区不存在。
\end{CJK}
\tcblower
Examples: \textbf{\{input\_content\}}

The above are some of the block names in Beijing, please imitate these block names and then make up at least 20 block names, and the requirements are not to duplicate the block names that exist in Beijing. Please double-check before exporting to make sure that the block name you are exporting does not exist.
\end{tcolorbox}

\subsection{Fictional Block Name Production (With Real Estate Developer)}

\begin{tcolorbox}[title = {Prompt}, breakable]
\begin{CJK}{UTF8}{gbsn}
参考样例：\textbf{\{input\_content\}}

以上是一些北京的小区名，请你模仿这些小区名再虚构至少20个小区名，要求在名字前面加上一些知名地产开发商（如“中海”“保利”“碧桂园”“恒大”等），且不能与北京现实中存在的小区名重复。输出前请再检查一遍确保你输出的小区不存在。
\end{CJK}
\tcblower
Examples: \textbf{\{input\_content\}}

The above are some of the block names in Beijing, please imitate these block names and then create at least 20 block names, and require some well-known real estate developers (such as "China Overseas", "Poly", "Country Garden", "Evergrande", etc.) in front of the name, and cannot be duplicated with the block names that exist in Beijing in reality. Please double-check before exporting to make sure that the block name you are exporting does not exist.
\end{tcolorbox}

\subsection{Fictional Block Name Production (Old block Style)}

\begin{tcolorbox}[title = {Prompt}, breakable]
\begin{CJK}{UTF8}{gbsn}
参考样例：\textbf{\{input\_content\}}

以上是一些北京的小区名，请你模仿这些小区名再虚构至少20个小区名，要求不要太过华丽，比较平淡、大众化，听起来像上个世纪的老旧小区，且不能与北京现实中存在的小区名重复。输出前请再检查一遍确保你输出的小区不存在。
\end{CJK}
\tcblower
Examples: \textbf{\{input\_content\}}

The above are some of the block names in Beijing, please imitate these block names and then make up at least 20 block names, the requirements are not too gorgeous, relatively plain, popular, it sounds like the old block of the last century, and can not be repeated with the block names that exist in Beijing in reality. Please double-check before exporting to make sure that the block name you are exporting does not exist.
\end{tcolorbox}

\subsection{Fictional Block Name Production (With Real Street or Town)}

\begin{tcolorbox}[title = {Prompt}, breakable]
\begin{CJK}{UTF8}{gbsn}
参考样例：\textbf{\{input\_content\}}

以上是一些北京的小区名，请你模仿这些小区名再虚构至少20个小区名，要求名字与北京的街道或乡镇名结合，且不能与北京现实中存在的小区名重复。输出前请再检查一遍确保你输出的小区不存在。
\end{CJK}
\tcblower
Example: \textbf{\{input\_content\}}

The above are some of the block names in Beijing, please imitate these block names and then create at least 20 block names, and the names are required to be combined with the names of streets or towns in Beijing, and cannot be duplicated with the block names that exist in Beijing in reality. Please double-check before exporting to make sure that the block name you are exporting does not exist.
\end{tcolorbox}

\subsection{Fictional Block Name Production (Institutional Dormitory Style)}

\begin{tcolorbox}[title = {Prompt}, breakable]
\begin{CJK}{UTF8}{gbsn}
参考样例：\textbf{\{input\_content\}}

以上是一些北京的小区名，请你模仿这些小区名再虚构至少20个小区名，要求名字是单位宿舍名，且不能与北京现实中存在的小区名重复。输出前请再检查一遍确保你输出的小区不存在。
\end{CJK}
\tcblower
Examples: \textbf{\{input\_content\}}

The above are some of the block names in Beijing, please imitate these block names and then create at least 20 block names, the name is required to be the name of the institutional dormitory, and can not be duplicated with the block name that exists in Beijing in reality. Please double-check before exporting to make sure that the cell you are exporting does not exist.
\end{tcolorbox}

\subsection{Fictional Area Name Production}

\begin{tcolorbox}[title = {Prompt}, breakable]
\begin{CJK}{UTF8}{gbsn}
参考样例：\textbf{\{input\_content\}}

以上是一些北京的地名，请你模仿这些区域名再虚构至少20个地名，且一定不能与北京现实中存在的地名重复。输出前请再检查一遍确保你输出的地名不存在。
\end{CJK}
\tcblower
Examples: \textbf{\{input\_content\}}

The above are some of the area names in Beijing, please imitate these area names and then create at least 20 area names, and must not duplicate the area names that exist in Beijing. Please double-check before exporting to make sure that the area name you are exporting does not exist.
\end{tcolorbox}

\section{\label{app:c} Prompts for Evaluation}

Appendix C shows the prompts used in evaluation experiments.

\subsection{Prompt Used by Test Model in  Ability Evaluation (With Context)}

\begin{tcolorbox}[title = {Prompt}, breakable]
\begin{CJK}{UTF8}{gbsn}
以下是中国关于房地产经纪人考试的单项选择题，考察内容为【\textbf{\{input\_category\}}】，你、客户、题目中出现的地点和小区均位于北京市。请你阅读上下文并根据题目要求从四个选项中选出最合适的选项，只输出正确选项对应的一个字母，不要输出其他内容。

上下文：

\textbf{\{input\_context\}}

题目：

\textbf{\{input\_question\}}

选项：

\textbf{\{input\_option\_a\}}

\textbf{\{input\_option\_b\}}

\textbf{\{input\_option\_c\}}

\textbf{\{input\_option\_d\}}

答案：
\end{CJK}
\tcblower
The following is a multiple-choice question about the real estate agent exam in China, which examines [\textbf{\{input\_category\}}], and you, the client, the location and block in which the question appears are all located in Beijing. Please read the context and choose the most appropriate option from the four options according to the requirements of the question, and only output the letter corresponding to the correct option, and nothing else.

Context:

\textbf{\{input\_context\}}

Question:

\textbf{\{input\_question\}}

Options:

\textbf{\{input\_option\_a\}}

\textbf{\{input\_option\_b\}}

\textbf{\{input\_option\_c\}}

\textbf{\{input\_option\_d\}}

Answer:
\end{tcolorbox}

\subsection{Prompt Used by Test Model in  Ability Evaluation (Without Context)}

\begin{tcolorbox}[title = {Prompt}, breakable]
\begin{CJK}{UTF8}{gbsn}
以下是中国关于房地产经纪人考试的单项选择题，考察内容为【\textbf{\{input\_category\}}】，你、客户、题目中出现的地点和小区均位于北京市。请你根据题目要求从四个选项中选出最合适的选项，只输出正确选项对应的一个字母，不要输出其他内容。

题目：

\textbf{\{input\_question\}}

选项：

\textbf{\{input\_option\_a\}}

\textbf{\{input\_option\_b\}}

\textbf{\{input\_option\_c\}}

\textbf{\{input\_option\_d\}}

答案：
\end{CJK}
\tcblower
The following is a multiple-choice question about the real estate agent exam in China, which examines [\textbf{\{input\_category\}}], and you, the client, the location and block in which the question appears are all located in Beijing. Please choose the most suitable option from the four options according to the requirements of the question, and only output one letter corresponding to the correct option, and do not output other content.

Question:

\textbf{\{input\_question\}}

Options:

\textbf{\{input\_option\_a\}}

\textbf{\{input\_option\_b\}}

\textbf{\{input\_option\_c\}}

\textbf{\{input\_option\_d\}}

Answer:
\end{tcolorbox}

\subsection{Prompt Used by Test Model in Hallucination Evaluation}

\begin{tcolorbox}[title = {Prompt}, breakable]
\begin{CJK}{UTF8}{gbsn}
你是一名北京市的专业的房地产经纪人，你只了解北京市的房产（如板块、商圈、楼盘、小区、小区周边设施等）信息。请你回答以下客户提问：

\textbf{\{input\_query\}}
\end{CJK}
\tcblower
You are a professional real estate agent in Beijing, and you only know the information of real estate in Beijing (such as plates, bizcircles, real estate, blocks, surrounding facilities, etc.). Please answer the following customer questions:

\textbf{\{input\_query\}}
\end{tcolorbox}

\subsection{Prompt Used by Judge Model in Hallucination Evaluation}

\begin{tcolorbox}[title = {Prompt}, breakable]
\begin{CJK}{UTF8}{gbsn}
前提：\textbf{\{input\_premise\}}

假设：\textbf{\{input\_hypothesis\}}

请你判断给定的“前提”和“假设”两段文本之间的关系属于以下哪一种：

1.蕴含：假设可以从前提中推断得到；

2.矛盾：假设与前提相矛盾；

3.中立：假设与前提既不矛盾也无法从前提中推断得到。

请用一个JSON输出你的检测结果，JSON包含"decision"（结论，有蕴含、矛盾或中立三种）和"explanation"（理由）两个字段，例如：

\{"decision": "蕴含", "explanation": "xxx（解释理由）"\}

\{"decision": "矛盾", "explanation": "xxx（解释理由）"\}

\{"decision": "中立", "explanation": "xxx（解释理由）"\}
\end{CJK}
\tcblower
Premise: \textbf{\{input\_premise\}}

Hypothesis: \textbf{\{input\_hypothesis\}}

Determine which of the following is the relationship between the given "premise" and "hypothesis" paragraphs:

1. Entailment: Hypotheses can be inferred from premises.

2. Contradiction: The hypothesis contradicts the premise.

3. Neutral: The hypothesis and the premise are neither contradictory nor can be inferred from the premise.

Please use a JSON to output your result, the JSON contains "decision" (conclusion, there are three types: Entailment, Contradiction, or Neutral) and "explanation", for example:

\{"decision": "Entailment", "explanation": "xxx"\}

\{"decision": "Contradiction", "explanation": "xxx"\}

\{"decision": "Neutral", "explanation": "xxx"\}
\end{tcolorbox}

\end{document}